\documentclass[sn-nature]{sn-jnl}

\usepackage{graphicx}%
\usepackage{multirow}%
\usepackage{amsmath,amssymb,amsfonts}%
\usepackage{amsthm}%
\usepackage{mathrsfs}%
\usepackage[title]{appendix}%
\usepackage{xcolor}%
\usepackage{textcomp}%
\usepackage{manyfoot}%
\usepackage{booktabs}%
\usepackage{algorithm}%
\usepackage{algorithmicx}%
\usepackage{algpseudocode}%
\usepackage{listings}%

\theoremstyle{thmstyleone}%
\theoremstyle{thmstyletwo}%

\theoremstyle{thmstylethree}%

\begin{document}

\title[Article Title]{Integrating AI in College Education:
Positive yet Mixed Experiences with ChatGPT}


\author[1]{\fnm{Xinrui} \sur{Song}}
\equalcont{These authors contributed equally to this work.}

\author[1]{\fnm{Jiajin} \sur{Zhang}}
\equalcont{These authors contributed equally to this work.}

\author[1]{\fnm{Pingkun} \sur{Yan}}
\author[1]{\fnm{Juergen} \sur{Hahn}}
\author[1]{\fnm{Uwe} \sur{Kruger}}

\author*[1]{\fnm{Hisham} \sur{Mohamed}}\email{mohamh2@rpi.edu}

\author*[1]{\fnm{Ge} \sur{Wang}}\email{wangg6@rpi.edu}

\affil*[1]{\orgdiv{Department of Biomedical Engineering}, \orgname{Rensselaer Polytechnic Institute}, \orgaddress{\street{1623 15th St}, \city{Troy}, \postcode{12180}, \state{NY}, \country{USA}}}


\abstract{The integration of artificial intelligence (AI) chatbots into higher education marks a shift towards a new generation of pedagogical tools, mirroring the arrival of milestones like the internet. With the launch of ChatGPT-4 Turbo in November 2023, we developed a ChatGPT-based teaching application (https://chat.openai.com/g/g-1imx1py4K-chatge-medical-imaging) and integrated it into our undergraduate medical imaging course in the Spring 2024 semester. This study investigates the use of ChatGPT throughout a semester-long trial, providing insights into students' engagement, perception, and the overall educational effectiveness of the technology. We systematically collected and analyzed data concerning students' interaction with ChatGPT, focusing on their attitudes, concerns, and usage patterns. The findings indicate that ChatGPT offers significant advantages such as improved information access and increased interactivity, but its adoption is accompanied by concerns about the accuracy of the information provided and the necessity for well-defined guidelines to optimize its use. }




\maketitle

\section{Introduction}\label{sec1}

Large language models (LLMs) represent a significant advancement in artificial intelligence (AI), based on their inherent model capacities and other cutting-edge deep learning technologies, such as self-supervised pre-training. The principle of next token prediction allows the generation of text or more generally a sequence of multimedia tokens that is not only coherent but also contextually relevant. Trained on extensive datasets, these deep models learn complex knowledge without explicit task-specific instructions. The versatility of LLMs and large vision-language models offers opportunities for artificial general intelligence and enables a broad spectrum of applications. 

Launched on November 6, 2023, ChatGPT-4 Turbo represents a major enhancement of the family of Generative Pre-trained Transformers (GPTs). This advanced AI model builds upon the capabilities of its predecessors by synergizing text, code and image data more broadly and more deeply. The model is optimized for real-time responses that are essential for time-sensitive applications such as interactive learning, produces high-quality and contextually appropriate responses, and in particular creates sophisticated AI-driven interactions in an educational scenario.

Based on the ChatGPT-4 Turbo platform, we developed a teaching application (https://chat.openai.com/g/g-1imx1py4K-chatge-medical-imaging) for our undergraduate medical imaging course in the Spring 2024 semester. We were motivated by both the human-like communication capability of ChatGPT and the observation that the younger generations, especially college students, quickly adapt to new technologies.

While higher education is generally cautious about the influences of IT innovations, our university has a long tradition in pedagogical innovation, allowing us to quickly investigate the use of AI in the classroom.

We believe that the potential impact of ChatGPT and other chatbots on education will be comparable to or even greater than the internet technology which has dramatically expanded the information sources available to college students. While higher education showed initial skepticism to this new technology\cite{baer1998will}, today it is difficult to imagine a university without a Learning Management System (LMS). Furthermore, the reliance on platforms like YouTube exemplifies this shift \cite{fleck2014youtube}. The popularity of mobile phones further facilitates access to these resources, a transition embraced by students \cite{al2016investigating}. The COVID-19 pandemic has underscored the necessity and merits of modern technological infrastructure.

While the internet brought various informational resources to students, ChatGPT-type models make retrieval and extraction of knowledge much more effective and efficient. These AI models have the potential to enable individualized learning when they are used as teaching assistants (TA) or tutors. However, even for ChatGPT-4 Turbo, the state-of-the-art AI-based chatbot, there remain uncertainties about what educational utilities it offers, how much value it adds, and how to address various associated challenges. Recently, commentaries raised concerns about the adverse consequences of ChatGPT's integration into educational settings \cite{milano2023large,fuchs2023exploring}. 

Given the above considerations, here we evaluate our ChatGPT-4 Turbo software through a semester-long trial. 
We systematically collected and analyzed data concerning students' interactions with ChatGPT. Also, we compared students' feedback on different ChatGPT-based chatbots within the same course.
In the next section, we describe our methods. Then, we summarize our key results and discuss relevant issues.

\section{Methods}\label{sec:methods}

In the study, three ChatGPT models are utilized with varying capabilities: 
(1) ChatGPT-3.5 serves as a freely accessible baseline, 
(2) ChatGPT-4 Turbo represents a significantly enhanced version, and
(3) a specialized application we developed, referred to as ChatGe-Medical-Imaging or ChatGe, is tailored for teaching a college-level medical imaging course.
To remove any financial barrier, we made a policy that any student can be reimbursed for one-month use of ChatGPT-4 Turbo.
In the Spring 2024 semester, two separate sections of the undergraduate medical imaging course were taught by two instructors Drs. Hisham Mohamed and Ge Wang respectively, with enrollments of 36 and 34 students. 
The two instructors shared the syllabus, the textbook, and exam questions, as well as three MATLAB sessions to provide students with hands-on experience.
According to the timeline given in the Supplementary Information (SI), we surveyed our students through three channels: students' questionnaires, course content Q\&A evaluations, and students' exam grades, and analyze the data as follows.

\section{Results and Discussions}\label{sec:resultAndDiscussion}

\subsection{Students' Survey Results on ChatGPT Usage and Assessment}
\label{sec:user exp}
\begin{figure*}[t!]
\centering
\includegraphics[width=.85\linewidth]{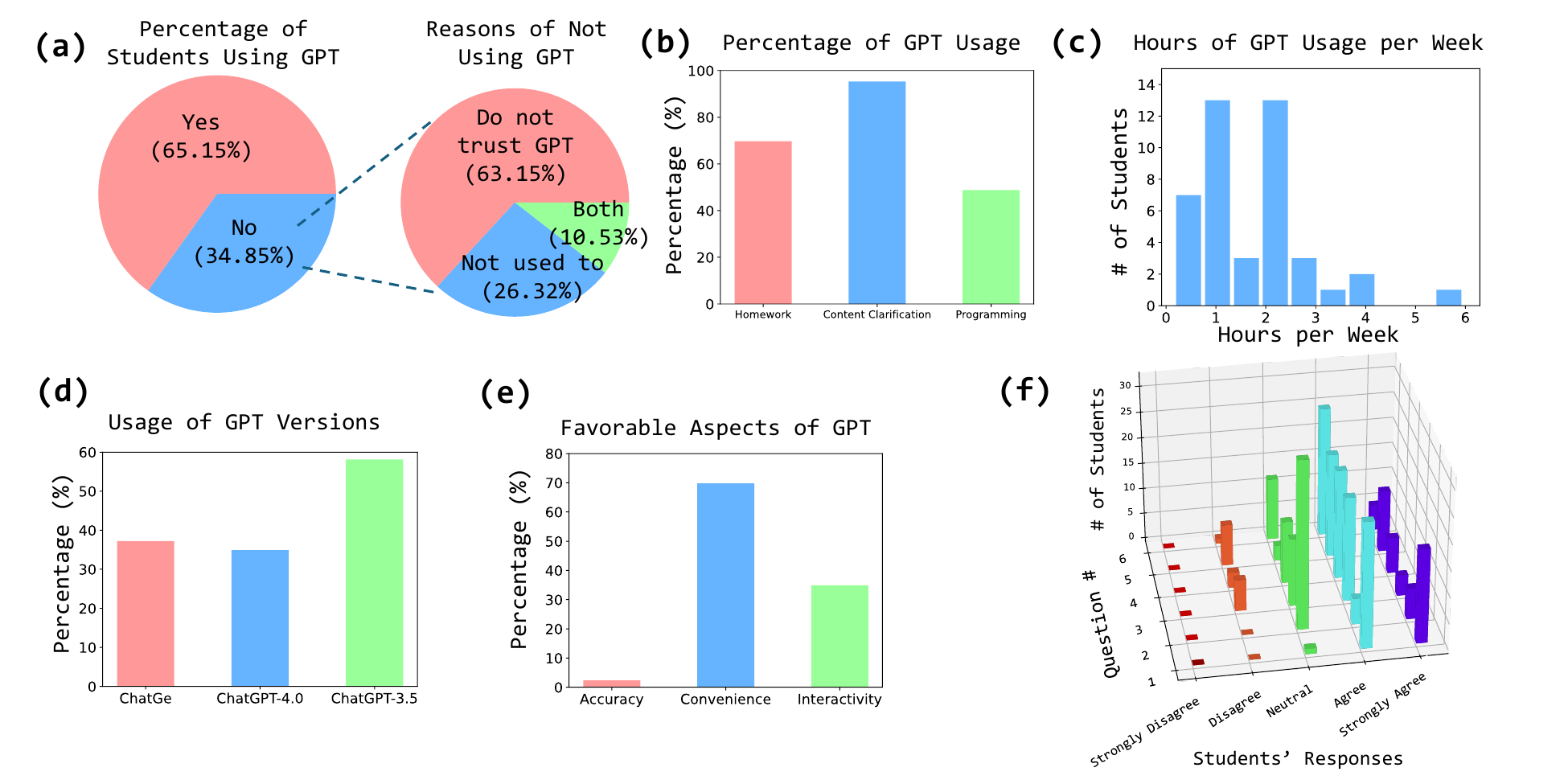}
\caption{Visualization of the students' feedback on AI-assisted course learning. The sub-figures \textbf{(a)}-\textbf{(e)} reflect the students' ChatGPT usage and impression. The sub-figure \textbf{(f} depicts the students' evaluation on various aspects of the chatbots. 
\textbf{(a)} The percentage of the students using ChatGPT in the two classes; \textbf{(b)} the use of ChatGPT; \textbf{(c)} the number of hours of the ChatGPT usage per week; \textbf{(d)} the usage of different chatbot versions; \textbf{(e)} the best aspects of these chatbots; and \textbf{(f)} the students' answers to six questions (see the SI for details) on various aspects of the chatbots. 
} 
\label{fig:user_exp}
\end{figure*}

Fig.~\ref{fig:user_exp} presents students' survey results from both sections on their ChatGPT usage and feedback. As shown in Fig.~\ref{fig:user_exp} \textbf{(a)}, about 65\% of the students from both sections used the ChatGPT models. Moreover, Fig.~\ref{fig:user_exp} \textbf{(b)} highlights that nearly all the users were aided by the chatbots for clarification of some course contents, and over half of them received ChatGPT's help for programming and homework. 
In addition to that, Fig.~\ref{fig:user_exp} \textbf{(c)} and \textbf{(d)} give information on how many hours per week the students spent using which chatbots.
Next, Fig.~\ref{fig:user_exp} \textbf{(e)} confirms that the students are most impressed with the convenience and interactivity of the ChatGPT models. However, few students felt highly confident in chatbots' responses. These data indicate that the students encountered false/untrustworthy answers by ChatGPT and/or they were not able to make good judgments. This is also supported by the students' reasons for not using ChatGPT, as shown in Fig.~\ref{fig:user_exp} \textbf{(a)}.
Finally, Fig.~\ref{fig:user_exp} \textbf{(f)} presents the students' answers to the six survey questions listed in 
the SI, which are related to different aspects of ChatGPT usage. The answers are required to be on a 5-point scale so that the values above 3 are positive while the values below 3 are negative. Evidently, the general attitude towards ChatGPT use in this context is overwhelmingly positive, except for Question \#2 which asks if students feel ChatGe-V1 significantly better than ChatGPT-4 Turbo. This negative feedback motivated us to upgrade ChatGe-V1 to ChatGe-V2 with more course-related materials and more specific prompts as further described in the SI. 

\subsection{Students' Preference over Different Versions of ChatGPT Tools}

As mentioned already, in our course we introduced the customized ChatGPT-4 Turbo app referred to as ChatGe, whose first version is ChatGe-V1, to assist students in understanding the complex theory of medical imaging. After each lecture, the students were asked to submit questions related to the new material. The instructors and TAs selected two most relevant questions (see the SI) and prepared answers using either a human TA, ChatGPT-4, or ChatGe-V1, and then selected two out of the three answers for students to indicate and explain their preference of one over the other. 

\begin{figure*}[t!]
\centering
\includegraphics[width=.75\linewidth]{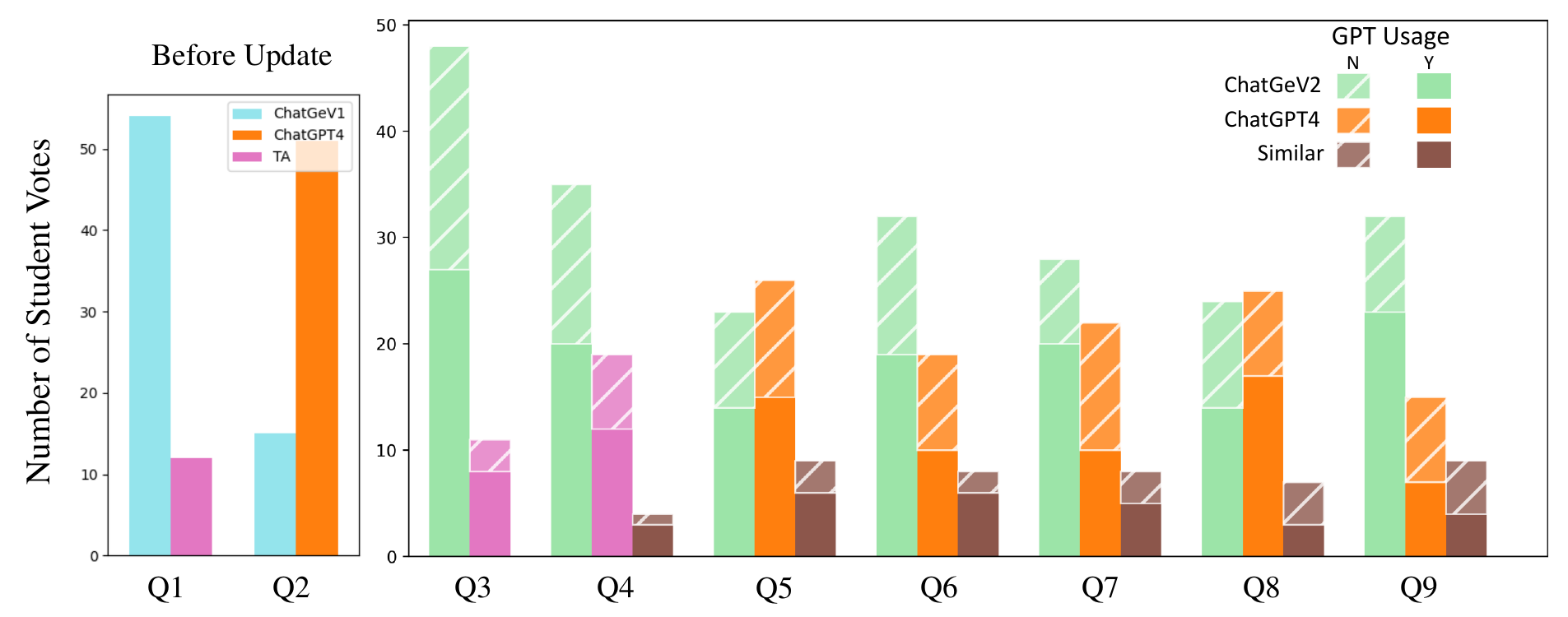}
\caption{Students' course content Q{\&}A preference. The bar plot statistically compares students' preference for the answers generated by either ChatGPT or TAs to typical course content-related questions. Q1 and Q2 were given after exam 1, and students responses were then collected to develop ChatGeV2.}
\label{fig:qa}
\end{figure*}

 Subsequently, we developed the second version of the app called ChatGe-V2, and repeated the comparative study on seven other questions. We asked ChatGPT-4 to summarize students' feedback on (Q1, Q2) to generate a prompt for ChatGe-V2 to enforce more tailored, student-friendly responses. ChatGe-V2 incorporates more course-relevant textbooks and other materials (more details on ChatGe-V1 and ChatGe-V2 can be found in the SI). Again, the students were asked to submit questions related to the new material, seven questions were selected (see the SI), and the answers were similarly generated for students to vote. Our analysis is summarized in Fig.\ref{fig:qa}, showing that while the original ChatGPT-4 performed better than ChatGe-V1, ChatGe-V2 improved the performance, being comparable or better than ChatGPT-4 Turbo, ChatGe-V1 and even human TAs. Interestingly, we found a statistically significant difference in the preference of students who have used ChatGPT for the course and those who have not. The students who used ChatGPT statistically prefer ChatGe-V2 (\( p\text{-value}=0.091 \text{ from Chi-squared test} \)), while no statistically significant preference was found among the rest. This indicates that students accustomed to reading ChatGPT-generated answers might be more capable of extracting information from the machine's responses. Overall, the students prefer more structured answers with bullet points. Most students dislike the upfront display of equations but some reported that formulas help when explained in detail. 

\subsection{Usage of ChatGPT Influenced by Promotions and Precaution}

Through the questionnaires, we found a significant difference in the students' adoption of ChatGPT, being strongly correlated to the instructor's promotional or pre-cautious emphases in the classroom. In a well-intended contrast,
one instructor actively promoted the use of ChatGPT, while the other more often cautioned the students about potential risks with ChatGPT. The impact of this contrast was evident in our post-exam survey: usage of ChatGPT was 87.5\% in one section where ChatGPT was encouraged, but only 38.2\% in the other section where ChatGPT was offered in a more conservative manner.
Among students who chose not to use ChatGPT, our data shows that over half of this population doubted the accuracy of ChatGPT (Fig. \ref{fig:user_exp} \textbf{(a)}). The remainder reported a lack of familiarity with the tools. As students become increasingly accustomed to chatbot assistants, the use of ChatGPT or similar tools is expected to rise. Moreover, advanced ChatGPT-type products now cite sources to support their output, alleviating the unreliability.

\subsection{ChatGPT for TA Workload Reduction}
As shown in Fig. \ref{fig:qa}, ChatGe-V2 generates answers much faster than our TAs (instantly vs. 15-20 minutes). Even when students' questions appear erroneous, ChatGe-V2 can infer the intended meaning of the question. Moreover, the TAs gave three lectures on MATLAB programming tasks. Instead of preparing the lectures directly, the TAs found that ChatGPT generated demonstrations as effectively as they can do. This example suggests that programming projects can also leverage the power of AI.

\section{Data and Code Availability Statement}
Data in this paper consists of survey results of each student. Anonymized data is available upon request. Survey questions are included in the supplementary material. All figures in this paper are generated with matplotlib in Python.

\section{Competing Interest}
The authors declare no competing interests.


\bibliography{reference}

\begin{thebibliography}{1}
\expandafter\ifx\csname url\endcsname\relax
  \def\url#1{\burl{#1}}\fi
\expandafter\ifx\csname urlprefix\endcsname\relax\def\urlprefix{URL }\fi
\providecommand{\bibinfo}[2]{#2}
\providecommand{\eprint}[2][]{\url{#2}}
\providecommand{\doi}[1]{\url{https://doi.org/#1}}
\bibcommenthead

\bibitem{baer1998will}
\bibinfo{author}{Baer, W.~S.}
\newblock \emph{\bibinfo{title}{Will the internet transform higher education?}}  (\bibinfo{publisher}{Rand Santa Monica, CA}, \bibinfo{year}{1998}).

\bibitem{fleck2014youtube}
\bibinfo{author}{Fleck, B.~K.}, \bibinfo{author}{Beckman, L.~M.}, \bibinfo{author}{Sterns, J.~L.} \& \bibinfo{author}{Hussey, H.~D.}
\newblock \bibinfo{title}{Youtube in the classroom: Helpful tips and student perceptions.}
\newblock \emph{\bibinfo{journal}{Journal of Effective Teaching}} \textbf{\bibinfo{volume}{14}}, \bibinfo{pages}{21--37} (\bibinfo{year}{2014}).

\bibitem{al2016investigating}
\bibinfo{author}{Al-Emran, M.}, \bibinfo{author}{Elsherif, H.~M.} \& \bibinfo{author}{Shaalan, K.}
\newblock \bibinfo{title}{Investigating attitudes towards the use of mobile learning in higher education}.
\newblock \emph{\bibinfo{journal}{Computers in Human behavior}} \textbf{\bibinfo{volume}{56}}, \bibinfo{pages}{93--102} (\bibinfo{year}{2016}).

\bibitem{milano2023large}
\bibinfo{author}{Milano, S.}, \bibinfo{author}{McGrane, J.~A.} \& \bibinfo{author}{Leonelli, S.}
\newblock \bibinfo{title}{Large language models challenge the future of higher education}.
\newblock \emph{\bibinfo{journal}{Nature Machine Intelligence}} \textbf{\bibinfo{volume}{5}}, \bibinfo{pages}{333--334} (\bibinfo{year}{2023}).

\bibitem{fuchs2023exploring}
\bibinfo{author}{Fuchs, K.}
\newblock \bibinfo{title}{Exploring the opportunities and challenges of nlp models in higher education: is chat gpt a blessing or a curse?} \textbf{\bibinfo{volume}{8}}, \bibinfo{pages}{1166682} (\bibinfo{year}{2023}).

\end{thebibliography}

\end{document}


\title[Article Title]{Supplementary Information}







\maketitle

\section{Materials for configuring and updating ChatGe}
\subsection{ChatGe-V1 Knowlege Base} The first version of our ChatGPT-4 Turbo based app is denoted as ChatGe-V1, which was developed in Dec., 2023 and made available to our students from the beginning of the course.  The ChatGe-V1 knowledge consists the following files:

\begin{enumerate}[leftmargin=1cm, itemsep=0pt, parsep=0pt]
\item[1.] Webb, Andrew. Introduction to Biomedical Imaging. IEEE Press, 2003.

\item[2.] Webb, Andrew. Solutions to Problems in Basic Principles of Medical Imaging. IEEE Press, 2006.

\item[3.] Wang, Ge. Basic Principles of Medical Imaging (draft).

\item[4.] Wang, Ge, and Mohamed, Hisham. Syllabus for BMED 2300 BioImaging and BioInstrumentation. Rensselaer Polytechnic Institute, 2019.

\item[5.] Past Homework Questions and Solutions, 2023
\end{enumerate}

\subsection{ChatGe-V2 Knowlege Base} The second version of our ChatGPT-4 Turbo based app is denoted as ChatGe-V2, which was upgraded in Feb., 2024 and made immediately available to our students to replace ChatGe-V1.  The ChatGe-V2 knowledge consists the following files:

\begin{enumerate}[leftmargin=1cm, itemsep=0pt, parsep=0pt]
\item[1.] Webb, Andrew. Introduction to Biomedical Imaging. IEEE Press, 2003.

\item[2.] Webb, Andrew. Solutions to Problems in Basic Principles of Medical Imaging. IEEE Press, 2006.

\item[3.] Wang, Ge. Basic Principles of Medical Imaging (draft).

\item[4.] Wang, Ge, and Mohamed, Hisham. Syllabus for BMED 2300 BioImaging and BioInstrumentation. Rensselaer Polytechnic Institute, 2019.

\item[5.] Past Homework Questions and Solutions, 2023

\item[6.] Bushberg, Jerrold T., Seibert, J. Anthony, Leidholdt, Edwin M. Jr., and Boone, John M. The Essential Physics of Medical Imaging. Lippincott Williams \& Wilkins, 3rd Edition.

\item[7.] Suetens, Paul. Fundamentals of Medical Imaging. Cambridge University Press, 2nd Edition.

\item[8.] Prince, Jerry L., and Links, Jonathan M. Medical Imaging Signals and Systems. Pearson Education, 2nd Edition.

\item[9.] Prince, Jerry L. Solutions Manual for Medical Imaging Signals and Systems. Pearson Education.

\item[10.] Maier, Andreas, Steidl, Stefan, Christlein, Vincent, and Hornegger, Joachim. Medical Imaging Systems: An Introductory Guide. Springer International Publishing.

\item[11.] Hendee, William R., and Ritenour, E. Russell. Medical Imaging Physics. Wiley-Liss, 4th Edition.

\item[12.] Wang G, Zhang Y, Ye XQ, Mou XQ: Machine Learning for Tomographic Imaging, IOP Publishing, 2019.
\end{enumerate}

Additionally, ChatGeV2 is guided with the following prompt generated by ChatGPT4 so that ChatGeV2 can answer questions in a manner that fits students' preferences:
"Please provide an answer that is concise and clear, incorporating practical examples or analogies to illustrate abstract concepts. Please structure the response with bullet points or sections for easy readability, ensuring it directly addresses the question being asked. Offer a balanced explanation that includes a brief overview for a quick grasp of the basic idea, followed by an option for more detailed information. Please include numerical examples or step-by-step breakdowns if applicable to help clarify complex or quantitative topics."

\newpage

\section{Opinion Survey Questions Cited in Figure 1}

\textbf{Fig. 1 (f)} summarizes the students' responses to the following six survey questions: 

\begin{enumerate}[leftmargin=1cm, itemsep=0pt, parsep=0pt]
    \item[1.] ChatGPT tools help learn the course and improve outcome;
    \item[2.] ChatGe-V1 is significantly better than ChatGPT-4 Turbo;
    \item[3.] ChatGPT tools contribute positively to the development of critical thinking and problem-solving skills; 
    \item[4.] ChatGPT tools help generate insights and ideas in learning the course contents;
    \item[5.] ChatGPT tools help personalize learning experience;
    \item[6.] the integration of ChatGPT tools positively affects the course.
\end{enumerate}

\newpage

\section{Course Content Questions Cited in Figure 2}

\textbf{Fig. 2} summarizes the students' responses to the following nine course content-related questions (Q1-Q9) selected from the pool of all the questions the students asked:
~\\

\noindent\textbf{Q1}: To clarify, is the Fourier Slice Theorem used to take a one-dimensional Fourier Transform and compare it to a two-dimensional Fourier Transform of the same function?
~\\

\noindent\textbf{Answer A}: Your description is not $100\%$ correct but captures some of the important components of the Fourier Slice Theorem.

\begingroup
\setlength{\parskip}{0pt} 
\begin{enumerate}[leftmargin=1cm, itemsep=0pt, parsep=0pt]
    \item[1.] Consider a 2D function $f(x,y)$ which could represent an image. Here, given coordinates x and y, f returns the intensity of a pixel. If we project this 2D function into a 1D space, creating a projection p(x) using a specific projection technique P, and then apply a Fourier transform to p(x), we obtain $F\{p(x)\}$.
    \item[2.] Similarly, if we apply a 2D Fourier transform directly to the 2D function f(x,y), we get $F\{f(x, y)\}$.
    \item[3.] By extracting a slice from $F\{f(x, y)\}$ along a line that is parallel to the direction of projection P, this slice is equivalent to $F\{p(x)\}$.
\end{enumerate}
\endgroup


In real-world applications, particularly in medical imaging, the 2D function f(x,y) can represent the true composition of the human body as seen in an axial plane. However, our imaging devices, which operate similarly to X-ray machines, are only capable of capturing a 1D projection of this plane. To reconstruct the axial view image, we collect multiple 1D projections of the plane at various angles. These projections can be assembled into a 2D Fourier image. By performing an inverse Fourier transform on this composite 2D image in Fourier space, we can transform it back into a 2D CT image that visually represents a plane within the human body.
~\\

\noindent\textbf{Answer B}: Your description is not 100\% correct but captures some of the important components of the Fourier Slice Theorem. It does not compare a one-dimensional Fourier Transform (1D FT) directly to a two-dimensional Fourier Transform (2D FT) of the same function in the way one might initially think. Instead, it establishes a relationship between the 1D Fourier Transform of a projection of a two-dimensional object and a slice through the origin of the 2D Fourier Transform of that object.
Here’s how it works in simpler terms:

\begingroup
\setlength{\parskip}{0pt} 
\setlist[enumerate]{nosep} 
\begin{enumerate}[leftmargin=1cm, itemsep=0pt, parsep=0pt]
    \item[1. ] Projection: Imagine you have a 2D object, and you shine light through it from one side to capture a shadow or projection on the other side. This projection is essentially a 2D object being represented in one dimension, where each point in the projection is a sum of the values along one line of the 2D object.
    \item[2. ] 1D Fourier Transform of the Projection: You take the Fourier Transform of this projection, which gives you a 1D Fourier representation. This process transforms the spatial information in the projection into frequency space, but it's still a 1D representation.
    \item[3. ] 2D Fourier Transform of the Object: Now, if you were to take the original 2D object and perform a 2D Fourier Transform on it, you would get a 2D representation in frequency space. This transform provides a comprehensive view of the spatial frequencies present in the 2D object.
    \item[4. ] Theorem Statement: The Fourier Slice Theorem states that the 1D Fourier Transform of the projection taken from a certain angle is equivalent to a slice, or a line, through the origin of the 2D Fourier Transform of the object, taken at the same angle. Essentially, it connects the process of projecting (a form of dimensionality reduction) with the process of slicing through the 2D frequency space representation of the original object.
\end{enumerate}
\endgroup

Hence, the theorem is not about comparing the transforms of the same function in 1D and 2D directly but is about relating the projection of an object (through its 1D Fourier Transform) with a slice of the 2D Fourier Transform of the object itself. This principle is crucial in reconstructing images from projections, as done in CT scans, where multiple projections at different angles are used to reconstruct a 2D image of the cross-section of an object or body.
~\\

\noindent\textbf{Your preference is \uline{\ \ \ \ } because \uline{\ \ \ \ \ \ \ \ \ \ \ \ \ \ \ \ \ \ \ \ \ \ \ \ \ \ \ \ \ \ }.}
~\\

\noindent\textcolor{gray}{\rule{\linewidth}{2pt}}
~\\

\noindent\textbf{Q2}: How many cross-sectional area photos/x rays is sufficient to complete an entire CT scan?
~\\

\noindent\textbf{Answer A}: The number of cross-sectional area photos or x-rays required to complete an entire CT (Computed Tomography) scan depends on several factors, including the specific anatomical area being imaged, the purpose of the scan, and the desired resolution. The fundamental principle of CT imaging involves taking multiple cross-sectional images (slices) of the body from different angles, which are then reconstructed into a three-dimensional image by a computer. This technique allows for the detailed visualization of internal structures.

Each cross-sectional image in a CT scan represents a "slice" of the body, similar to how a loaf of bread can be sliced to reveal each individual piece. The thickness of these slices can vary depending on the type of CT scan and the imaging requirements. For instance, high-resolution scans require thinner slices (e.g., 0.5 mm to 1 mm), while standard scans may use thicker slices (e.g., 5 mm to 10 mm).

The total number of slices needed for a complete CT scan is determined by dividing the total length of the area being scanned by the thickness of each slice. For example, if the area of interest is 150 mm long and the slice thickness is set to 5 mm, then the scan would require 30 slices (150 mm / 5 mm per slice = 30 slices).

In practice, the exact number of slices for a complete CT scan can vary widely. A scan focusing on a small, specific area, such as the inner ear, might require only a few dozen high-resolution slices, while a scan covering a larger region, such as the abdomen and chest, could require hundreds of slices, especially if high detail is needed.

It's also worth noting that modern CT scanners and imaging techniques continue to evolve, with some newer machines capable of capturing thinner slices and more detailed images, potentially affecting the total number of slices needed for a comprehensive scan.

Therefore, there isn't a one-size-fits-all answer to the number of cross-sectional images required for a CT scan; it's highly dependent on the scan's specific clinical requirements and the technology being used.
~\\

\noindent\textbf{Answer B}: The number of cross-sectional area images, or slices, needed to complete an entire CT (Computed Tomography) scan varies widely based on the purpose of the scan, the specific area of the body being scanned, and the detail required for diagnosis or treatment planning. CT scans work by taking multiple X-ray measurements from different angles, creating cross-sectional images of the body. These images can then be assembled to construct a 3D representation of the scanned area.

\begingroup
\setlength{\parskip}{0pt} 
\setlist[enumerate]{nosep} 
\begin{enumerate}[leftmargin=1cm, itemsep=0pt, parsep=0pt]
\item[1. ] \textbf{Scan Resolution:} Higher resolution scans require more slices. For instance, if a high level of detail is needed, such as in a brain scan to identify small lesions, more slices will be taken, possibly in the range of hundreds. High-resolution scans can have slice thicknesses as small as 0.5 mm.
\item[2. ] \textbf{Scanned Body Area:} The size of the area being scanned also impacts the number of slices. A scan of a smaller area, like the head, might need fewer slices compared to a scan of the entire chest, abdomen, and pelvis.
\item[3. ] \textbf{Scan Purpose:} The purpose of the scan can also influence the number of slices. For example, a CT scan looking for fractures might not need as many slices as one searching for small tumors or vascular abnormalities.
\item[4. ] \textbf{Technology and Protocol Variations:} Advances in CT technology and differences in scanning protocols between medical facilities can also affect the number of slices. Some modern CT scanners can take more slices in a shorter amount of time, thanks to improvements in speed and resolution.
\end{enumerate}
\endgroup

In general, a CT scan can range from a few dozen to several hundred slices. For a full-body scan, the number could be in the higher end of this range, especially if detailed images are needed. It's important to note that medical professionals determine the optimal number of slices based on the specific clinical situation, balancing the need for detailed images with the desire to minimize radiation exposure.
~\\

\noindent\textbf{Your preference is \uline{\ \ \ \ } because \uline{\ \ \ \ \ \ \ \ \ \ \ \ \ \ \ \ \ \ \ \ \ \ \ \ \ \ \ \ \ \ }.}
~\\

\noindent\textcolor{gray}{\rule{\linewidth}{2pt}}
~\\

\noindent\textbf{Q3}: How does positron emission differ from different types of decay?
~\\

\noindent\textbf{Answer A}: Positron emission is a type of radioactive decay that is unique due to the particles it emits and the way it interacts with matter. Here's how it differs from other common types of radioactive decay:

\begingroup
\setlength{\parskip}{0pt} 
\setlist[enumerate]{nosep} 
\begin{enumerate}[leftmargin=1cm, label=$\bullet$, itemsep=0pt, parsep=0pt]
\item \textbf{Alpha Decay:} In alpha decay, an alpha particle (which is essentially a helium nucleus consisting of two protons and two neutrons) is emitted from the nucleus of a radioactive atom. This results in a decrease in the atomic number by two and mass number by four, thus transforming the element into another.
\item \textbf{Beta Decay:} Beta decay involves the transformation of a neutron into a proton within the nucleus, accompanied by the emission of a beta particle (electron) and an anti-neutrino. In beta-minus decay, a neutron is converted to a proton, increasing the atomic number by one. There is also beta-plus decay, where a proton is converted into a neutron and a positron (the antimatter counterpart of an electron) and a neutrino are emitted.
\item \textbf{Gamma Decay:} Gamma decay occurs after other types of decay, where the nucleus remains in an excited state with excess energy. This excess energy is released in the form of gamma rays, high-energy photons, without changing the number of protons or neutrons in the nucleus.
\item \textbf{Positron Emission:} Specifically, positron emission is a form of beta-plus decay. Here, a proton in the nucleus is transformed into a neutron, releasing a positron and a neutrino. The emitted positron will travel until it encounters an electron, at which point both particles annihilate each other, typically resulting in the production of two or more gamma photons. This annihilation is distinctive and forms the basis for Positron Emission Tomography (PET) scans used in medical imaging, allowing for detailed metabolic and physiologic imaging.
\end{enumerate}
\endgroup

The main differences lie in the particles emitted (positron vs. electron or alpha particles) and the resultant transformations within the atomic nucleus. Positron emission is particularly notable for its application in medical imaging, contrasting with the primarily analytical or energy considerations of other decay types.
~\\

\noindent\textbf{Answer B}: Radioactive decay can be broadly categorized into Alpha Decay and Beta Decay. Alpha Decay involves the emission of alpha particles, which consist of two protons and two neutrons. In contrast, Beta Decay involves the emission of beta particles, which are high-energy, high-speed electrons ($\beta$-) or positrons ($\beta$+). This leads to further subdivision of Beta Decay into Beta-minus decay, involving electrons, and Beta-plus decay, involving positrons—the latter being essential to Positron Emission Tomography (PET) imaging.

Additionally, we have discussed Gamma rays, which are fundamental to Single Photon Emission Computed Tomography (SPECT). Unlike Alpha and Beta particles, Gamma rays are massless photons that transmit energy. Gamma rays are often emitted alongside Alpha and Beta decays or when a nucleus transitions from an excited state back to its ground state.
~\\

\noindent\textbf{Your preference is \uline{\ \ \ \ } because \uline{\ \ \ \ \ \ \ \ \ \ \ \ \ \ \ \ \ \ \ \ \ \ \ \ \ \ \ \ \ \ }.}
~\\

\noindent\textcolor{gray}{\rule{\linewidth}{2pt}}
~\\

\newpage

\noindent\textbf{Q4}: How does the Point Spread Function (PSF) relate to image resolution in medical imaging systems, and what is the significance of the Gaussian tail?
~\\

\noindent\textbf{Answer A}:
\begingroup
\setlength{\parskip}{0pt} 
\setlist[enumerate]{nosep} 
\begin{enumerate}[leftmargin=1cm, label=$\bullet$, itemsep=0pt, parsep=0pt]
\item[1. ] \textbf{Understanding the Point Spread Function (PSF)} The PSF describes how an imaging system responds to a point source, influencing image quality and resolution. Essentially, it indicates how a system blurs or spreads out light or radiation. A narrow PSF enhances the system's ability to resolve fine details, improving resolution.
\item[2. ]  \textbf{PSF and Image Resolution} Image resolution depends directly on the PSF. A narrow PSF enables the system to distinguish clearly between closely spaced objects, improving resolution. Conversely, a wider PSF blends objects together, reducing resolution. The minimum distance at which two points can be resolved—often guided by the Rayleigh criterion—is determined by the PSF's width.
\item[3. ] \textbf{Gaussian PSF and Its Gaussian Tail} Often, the PSF can be modeled as a Gaussian function, noted for its bell-shaped curve. The central peak's height and sharpness affect the image's detail and contrast, with a sharper peak providing better resolution.
\end{enumerate}
\endgroup

Practical implication of managing the Gaussian tail is critical for:
\begingroup
\setlength{\parskip}{0pt} 
\setlist[enumerate]{nosep} 
\begin{enumerate}[leftmargin=1cm, label=$\bullet$, itemsep=0pt, parsep=0pt]
\item Enhancing Image Clarity: Narrowing the Gaussian tail reduces signal spread, enhancing clarity and contrast.
\item Reducing Artifacts: A shorter tail decreases overlap between structures, reducing artifacts and improving accuracy.
\item Optimizing System Design: Achieving an optimal PSF involves balancing resolution with the effects of physical processes like scattering, which can widen the PSF.
Understanding and controlling the PSF, especially the characteristics of its Gaussian tail, is vital for improving image quality and diagnostic capabilities in medical imaging systems.
\end{enumerate}
\endgroup
~\\

\noindent\textbf{Answer B}: Here’s a condensed explanation of how the Point Spread Function (PSF) relates to image resolution in medical imaging systems, emphasizing the importance of the Gaussian tail:

Overview of PSF and Image Resolution:
\begingroup
\setlength{\parskip}{0pt} 
\setlist[enumerate]{nosep} 
\begin{enumerate}[leftmargin=1cm, label=$\bullet$, itemsep=0pt, parsep=0pt]
\item \textbf{Point Spread Function (PSF):} Describes the imaging system's response to a point source, crucial for determining how details are depicted in an image.
\item \textbf{Image Resolution:} Relates to the system's ability to clearly differentiate close objects, heavily influenced by the sharpness of the PSF.
Relationship Between PSF and Image Resolution
\item \textbf{Sharper PSF Equals Higher Resolution:} A narrow PSF results in sharper images, enhancing the system’s resolution by accurately depicting small details.
\item \textbf{Wider PSF Leads to Blurring:} A broad PSF causes blurring, reducing the resolution as details merge together, making them harder to distinguish.
Significance of the Gaussian Tail in PSF
\item \textbf{Characteristics of Gaussian PSF:} A Gaussian PSF, common in MRI and CT scans, features a central peak and a tapering tail that influences edge clarity.
\item \textbf{Trade-off in Design:} A longer Gaussian tail can improve the signal-to-noise ratio (SNR) by capturing more signal from the point source, albeit at some loss of resolution.
\item \textbf{Optimization and Deconvolution:} System designers optimize the PSF to balance resolution and noise. Techniques like deconvolution are used post-acquisition to refine the PSF, aiming to improve resolution further.
Overall, the PSF’s shape, particularly the extent of its Gaussian tail, plays a pivotal role in defining the resolution and image quality in medical imaging, guiding system design and image processing strategies to produce optimal diagnostic images.
\end{enumerate}
\endgroup

\noindent\textbf{Your preference is \uline{\ \ \ \ } because \uline{\ \ \ \ \ \ \ \ \ \ \ \ \ \ \ \ \ \ \ \ \ \ \ \ \ \ \ \ \ \ }.}
~\\

\noindent\textcolor{gray}{\rule{\linewidth}{2pt}}
~\\

\noindent\textbf{Q5}: How does the use of mechanical collimators in SPECT imaging compare to the 3D electronic collimation employed in PET imaging in terms of sensitivity and specificity of detecting abnormalities?
~\\

\noindent\textbf{Answer A}: The use of mechanical collimators in Single Photon Emission Computed Tomography (SPECT) and 3D electronic collimation in Positron Emission Tomography (PET) imaging have distinct impacts on their sensitivity and specificity in detecting abnormalities.

\begingroup
\setlength{\parskip}{0pt} 
\setlist[enumerate]{nosep} 
\noindent \textbf{SPECT Imaging with Mechanical Collimators}
\begin{enumerate}[leftmargin=1cm, label=$\bullet$, itemsep=0pt, parsep=0pt]
\item \textbf{Mechanical Collimation}: SPECT uses physical devices that filter gamma rays based on direction, impacting the types of images produced.
\item \textbf{Sensitivity}: SPECT's sensitivity is relatively lower because mechanical collimators block many incoming rays, requiring higher doses or longer imaging times for adequate results.
\item \textbf{Specificity}: While SPECT can achieve good specificity with the correct collimator, its resolution is inherently limited by the physical design of these devices.
\end{enumerate}
\endgroup

\noindent \textbf{PET Imaging with 3D Electronic Collimation}
\begingroup
\setlength{\parskip}{0pt} 
\setlist[enumerate]{nosep} 
\begin{enumerate}[leftmargin=1cm, label=$\bullet$, itemsep=0pt, parsep=0pt]
\item \textbf{Electronic Collimation}: PET employs coincidence detection, where simultaneous detection of two photons indicates the emission point, enhancing imaging efficiency.
\item \textbf{Sensitivity}: PET is generally more sensitive than SPECT as it captures all emitted photons that meet the coincidence criteria, increasing detectable events.
\item \textbf{Specificity}: The high sensitivity and advanced image processing in PET contribute to superior spatial resolution and specificity, helping better identify and differentiate abnormalities.
Comparative Analysis
\item \textbf{Quantitative Capability}: PET provides quantitative measurements of biochemical changes, improving both sensitivity and specificity.
\item \textbf{Resolution}: PET typically offers better resolution due to its more efficient photon utilization and absence of physical collimators.
\item \textbf{Radioisotope Utilization}: PET's broader range of isotopes and efficient detection enhance its diagnostic effectiveness compared to SPECT.
\end{enumerate}
\endgroup

In summary, PET imaging's 3D electronic collimation generally outperforms SPECT's mechanical collimators in sensitivity and specificity, making it more suitable for a wider range of diagnostic applications, particularly in detecting molecular and functional abnormalities.
~\\

\noindent\textbf{Answer B}: Single Photon Emission Computed Tomography (SPECT) and Positron Emission Tomography (PET) are both nuclear imaging techniques that provide metabolic and functional information. Each utilizes different types of collimation techniques to form images from emitted radiation, impacting their sensitivity and specificity in detecting abnormalities. Here’s how mechanical and electronic collimation in SPECT and PET, respectively, compare:
\\

\begingroup
\setlength{\parskip}{0pt} 
\setlist[enumerate]{nosep} 
\noindent \textbf{Mechanical Collimation in SPECT}
\begin{enumerate}[leftmargin=1cm, label=$\bullet$, itemsep=0pt, parsep=0pt]
\item \textbf{Mechanical Collimators}: SPECT uses mechanical collimators (typically made of lead) to direct the gamma rays emitted from the patient towards the detector. These collimators only allow gamma rays traveling in specific directions to pass through, blocking others.
\item \textbf{Sensitivity}: The use of mechanical collimation in SPECT limits the system's sensitivity because many emitted photons are blocked and do not contribute to image formation. This leads to longer acquisition times to gather enough data for image reconstruction.
\item \textbf{Specificity}: Mechanical collimators improve the specificity by reducing the scatter and background radiation, enhancing image clarity and accuracy in identifying the location of radiotracers.
\end{enumerate}
\endgroup

\noindent \textbf{Electronic Collimation in PET}
\begingroup
\setlength{\parskip}{0pt} 
\setlist[enumerate]{nosep} 
\begin{enumerate}[leftmargin=1cm, label=$\bullet$, itemsep=0pt, parsep=0pt]
\item \textbf{Electronic Collimators}: PET uses electronic collimation, which involves detecting coincident photon pairs emitted in opposite directions from positron annihilations. This technique does not physically block any photons but identifies their paths electronically.
\item \textbf{Sensitivity}: PET is generally more sensitive than SPECT. The absence of physical collimators allows more photons to be detected, contributing to faster data acquisition and higher resolution images.
\item \textbf{Specificity}: The coincident detection inherent in PET imaging significantly reduces the background noise and increases specificity. This method is highly effective in identifying small and deep-seated abnormalities due to its ability to precisely localize the origin of the photon pairs.
\end{enumerate}
\endgroup

\noindent \textbf{Summary}
\begingroup
\setlength{\parskip}{0pt} 
\setlist[enumerate]{nosep} 
\begin{enumerate}[leftmargin=1cm, label=$\bullet$, itemsep=0pt, parsep=0pt]
\item \textbf{SPECT Imaging}: Higher specificity due to mechanical collimation but lower sensitivity as many photons are blocked.
\item \textbf{PET Imaging}: Higher sensitivity due to the use of electronic collimation allowing more photon detection, and higher specificity in detecting abnormalities through coincident photon pair detection.
Overall, PET imaging tends to provide higher sensitivity and specificity compared to SPECT due to its advanced collimation technique, which effectively increases the quality and diagnostic usefulness of the images.
\end{enumerate}
\endgroup

\noindent\textbf{Your preference is \uline{\ \ \ \ } because \uline{\ \ \ \ \ \ \ \ \ \ \ \ \ \ \ \ \ \ \ \ \ \ \ \ \ \ \ \ \ \ }.}
~\\

\noindent\textcolor{gray}{\rule{\linewidth}{2pt}}
~\\

\noindent\textbf{Q6}: What are the advantages of using a spiral imaging technique over conventional imaging methods in MRI?
~\\

\noindent\textbf{Answer A}: Spiral imaging techniques in MRI have several advantages over traditional Cartesian methods, primarily due to their high efficiency and effectiveness in minimizing certain types of artifacts:
\begingroup
\setlength{\parskip}{0pt} 
\setlist[enumerate]{nosep} 
\begin{enumerate}[leftmargin=1cm, label=$\bullet$, itemsep=0pt, parsep=0pt]
\item \textbf{Faster Data Acquisition}: Spiral imaging typically requires fewer passes through k-space to collect the necessary data, which shortens scan times. Unlike Cartesian sampling, which acquires data line-by-line in a stepwise fashion, spiral imaging captures data in a continuous motion. This approach is generally more efficient at filling k-space.
\item \textbf{Reduced Motion Artifacts}: Spiral imaging quickly acquires the core of k-space in one continuous sweep, making it more resistant to motion artifacts. This feature is particularly advantageous in clinical settings where patient movement is common.
\item \textbf{Improved Temporal Resolution}: The rapid data acquisition of spiral imaging enhances temporal resolution, making it particularly suitable for functional MRI (fMRI) and perfusion studies, where monitoring changes over time is essential.
\item \textbf{Fewer Artifacts}: The continuous nature of spiral data acquisition helps to inherently compensate for flow and motion, thus reducing artifacts associated with blood flow, which are often an issue in Cartesian scans.
\end{enumerate}
\endgroup

\noindent\textbf{Answer B}: Spiral (or helical) imaging techniques in MRI offer several advantages over conventional imaging methods, such as those using a Cartesian acquisition scheme. Here are some key benefits:
\begingroup
\setlength{\parskip}{0pt} 
\setlist[enumerate]{nosep} 
\begin{enumerate}[leftmargin=1cm, label=$\bullet$, itemsep=0pt, parsep=0pt]
\item \textbf{Increased Speed}: One of the most significant advantages of spiral imaging is its faster data acquisition. Unlike conventional methods that collect data row by row or slice by slice, spiral imaging continuously collects data in a spiral trajectory, starting at the center of k-space and moving outward. This allows for rapid coverage of k-space, which can be particularly beneficial for capturing dynamic processes like perfusion or for reducing motion artifacts from patient movement.
\item \textbf{Efficiency in Filling K-Space}: Spiral imaging is highly efficient in filling k-space because there are no gaps between lines as can happen with rectangular scanning patterns. This efficiency often results in better utilization of the magnetic fields and gradients.
\item \textbf{Reduced Motion Artifacts}: Due to the rapid data acquisition, spiral imaging is less susceptible to motion artifacts. This is particularly important for clinical applications involving the abdomen or thoracic cavity, where patient movement can be unpredictable.
\item \textbf{Flexibility in Image Contrast}: Spiral imaging provides flexibility in adjusting image contrast. The method allows for adjustments in the trajectory through k-space, which can be optimized to enhance certain types of image contrast or to suppress specific signal components, enhancing the diagnostic quality of the images.
\item \textbf{Potential for High-Resolution Imaging}: The continuous nature of the spiral trajectory allows for very high spatial resolution imaging. Because the data is acquired continuously, spiral scans can achieve higher resolution in a shorter time compared to conventional step-by-step methods.
\end{enumerate}
\endgroup

These advantages make spiral MRI particularly useful in situations where speed and motion sensitivity are crucial, and where high-resolution imaging is needed rapidly. This makes it valuable for clinical scenarios involving moving organs or for patients who cannot hold still for long periods.
~\\

\noindent\textbf{Your preference is \uline{\ \ \ \ } because \uline{\ \ \ \ \ \ \ \ \ \ \ \ \ \ \ \ \ \ \ \ \ \ \ \ \ \ \ \ \ \ }.}
~\\

\noindent\textcolor{gray}{\rule{\linewidth}{2pt}}
~\\

\noindent\textbf{Q7}: How does the frequency of the RF field relate to the Larmor frequency of the protons, and what would be the outcome if the frequency of the RF field is significantly different from the Larmor frequency?
~\\

\noindent\textbf{Answer A}: 

\textbf{(1) Relationship Between RF Frequency and Larmor Frequency}
\begingroup
\setlength{\parskip}{0pt} 
\setlist[enumerate]{nosep} 
\begin{enumerate}[leftmargin=1cm, label=$\bullet$, itemsep=0pt, parsep=0pt]
    \item \textbf{Larmor Frequency}: This is the frequency at which the protons in a magnetic field precess. It is determined by the gyromagnetic ratio of the proton and the strength of the magnetic field according to the equation \(\omega_0 = \gamma B_0\), where \(\omega_0\) is the Larmor frequency, \(\gamma\) is the gyromagnetic ratio, and \(B_0\) is the magnetic field strength.

	\item \textbf{RF Field Frequency}: For effective resonance and signal generation in MRI, the frequency of the RF field applied must match the Larmor frequency of the protons. When the RF field’s frequency matches the Larmor frequency, the RF field can efficiently exchange energy with the protons, flipping their spins from the lower energy state to the higher one, which is key for MRI signal generation.
\end{enumerate}
\endgroup

\textbf{(2) Outcome if the RF Frequency Diverges from the Larmor Frequency}
\begingroup
\setlength{\parskip}{0pt} 
\setlist[enumerate]{nosep} 
\begin{enumerate}[leftmargin=1cm, label=$\bullet$, itemsep=0pt, parsep=0pt]
	\item \textbf{Off-Resonance Effects}: If the frequency of the RF field is not equal to the Larmor frequency, the energy transfer between the RF field and the protons becomes inefficient. This condition is known as being off-resonance.
	\item \textbf{Reduced Signal Strength}: The most immediate effect of an off-resonance condition is a reduction in the strength of the MRI signal because fewer protons are effectively flipped to the higher energy state.
	\item \textbf{Image Artifacts}: Off-resonance can also lead to image artifacts. Protons in different parts of the tissue might experience slightly different magnetic fields due to inhomogeneities in the main magnetic field. If the RF frequency does not match the local Larmor frequency exactly, parts of the image may appear distorted or blurred.
\end{enumerate}
\endgroup

\textbf{(3) Practical Considerations}
\begingroup
\setlength{\parskip}{0pt} 
\setlist[enumerate]{nosep} 
\begin{enumerate}[leftmargin=1cm, label=$\bullet$, itemsep=0pt, parsep=0pt]
	\item \textbf{Shimming and Frequency Adjustment}: MRI systems often include procedures like shimming to homogenize the magnetic field and adjustments of the RF frequency to match the local Larmor frequency as closely as possible, minimizing off-resonance effects and improving image quality.
\end{enumerate}
\endgroup

Overall, precise matching of the RF field frequency to the Larmor frequency is essential for optimal MRI performance, affecting everything from signal strength to image clarity and accuracy.
~\\

\noindent\textbf{Answer B}: In magnetic resonance imaging (MRI) and nuclear magnetic resonance (NMR) spectroscopy, the relationship between the frequency of the radiofrequency (RF) field and the Larmor frequency of the protons is crucial.

The Larmor frequency is the frequency at which protons (or other nuclei) in a magnetic field precess (spin around the axis of the magnetic field). This frequency is directly proportional to the strength of the magnetic field and is given by the equation: \(\omega = \gamma B_0\) where:
\begingroup
\setlength{\parskip}{0pt} 
\setlist[itemize]{nosep} 
\begin{itemize}
	\item \(\omega\) is the Larmor frequency,
	\item \(\gamma\) is the gyromagnetic ratio (a constant specific to each type of nucleus),
	\item \(B_0\) is the magnetic field strength.
\end{itemize}
\endgroup

The RF field used in MRI or NMR must match the Larmor frequency of the protons (or other nuclei being observed). This frequency matching is important for a phenomenon called resonance. When the frequency of the RF field is equal to the Larmor frequency, it can effectively transfer energy to the protons, causing them to flip their spins from a lower energy state to a higher energy state (or vice versa). This is crucial for creating the signal that is detected and converted into images or spectroscopic data.

If the frequency of the RF field is significantly different from the Larmor frequency:
\begingroup
\setlength{\parskip}{0pt} 
\setlist[enumerate]{nosep} 
\begin{enumerate}[leftmargin=1cm, label=$\bullet$, itemsep=0pt, parsep=0pt]
	\item \textbf{Reduced Signal}: There would be a significant reduction in the efficiency of the energy transfer, leading to a weaker signal. This is because the RF field would not be in resonance with the spinning protons, so it would not effectively cause the spins to flip.
	\item \textbf{Poor Image Quality or Spectral Data}: In the context of MRI, this would result in poor image quality as the necessary contrast and details would not be properly captured. In NMR spectroscopy, the spectral peaks would be less distinct or could be completely absent, complicating the analysis of molecular structures.
	\item \textbf{No Resonance}: At a certain point, if the mismatch is too large, there might be no observable effect on the spins at all, rendering the RF field essentially useless for the purpose of MRI or NMR.
\end{enumerate}
\endgroup

Therefore, precise control of the RF frequency to match the Larmor frequency is essential for effective MRI and NMR procedures.
~\\

\noindent\textbf{Your preference is \uline{\ \ \ \ } because \uline{\ \ \ \ \ \ \ \ \ \ \ \ \ \ \ \ \ \ \ \ \ \ \ \ \ \ \ \ \ \ }.}
~\\

\noindent\textcolor{gray}{\rule{\linewidth}{2pt}}
~\\

\noindent\textbf{Q8}: Why do we need coupling media for ultrasound imaging?
~\\

\noindent\textbf{Answer A}: Coupling media, commonly referred to as ultrasound gel, is crucial for effective ultrasound imaging due to several key reasons:

\begingroup
\setlength{\parskip}{0pt} 
\setlist[enumerate]{nosep} 
\begin{enumerate}[leftmargin=1cm, label=$\bullet$, itemsep=0pt, parsep=0pt]
\item \textbf{Sound Wave Transmission}: Ultrasound waves are sound waves that travel through different mediums at varying speeds. In air, the transmission of these sound waves is poor because air is a much less dense medium compared to human tissues. The coupling medium acts as a conductive gel that fills the air gaps between the ultrasound probe and the skin, ensuring that the ultrasound waves pass efficiently into the body without being scattered or reflected by the air.
\item \textbf{Image Quality Enhancement}: The coupling medium helps in reducing the amount of air between the probe and the skin, which if present, can cause reflections and scattering of the ultrasound waves. This scattering can degrade the quality of the image captured. By using a coupling gel, the clarity and resolution of the ultrasound images are significantly improved.
\item \textbf{Reduced Friction and Comfort}: The gel helps in smoothly gliding the ultrasound probe over the skin surface. This not only makes the procedure more comfortable for the patient but also allows the technician to maneuver the probe more easily, facilitating better imaging and diagnostics.
\item \textbf{Patient Safety}: Using a coupling medium also helps in minimizing the heat generated by the friction between the probe and the skin. This is important for the comfort and safety of the patient during the ultrasound examination.
\end{enumerate}
\endgroup

Overall, the use of a coupling medium is essential for effective, safe, and high-quality ultrasound imaging. It ensures that ultrasound waves are transmitted efficiently into the body, which is crucial for accurate diagnostics.
~\\

\noindent\textbf{Answer B}: In ultrasound imaging, coupling media, commonly referred to as ultrasound gel, are crucial for effective image capture. Here’s a detailed explanation of why they are needed:

\noindent \textbf{Overview}
\begingroup
\setlength{\parskip}{0pt} 
\setlist[enumerate]{nosep} 
\begin{enumerate}[leftmargin=1cm, label=$\bullet$, itemsep=0pt, parsep=0pt]
\item \textbf{Purpose of Coupling Media}: Coupling media, such as gels, creams, or oils, are used to facilitate the transmission of ultrasound waves from the transducer into the body and back.
\item \textbf{Elimination of Air Gap}: The primary purpose is to eliminate any air gaps between the ultrasound transducer and the skin. Air is a poor conductor of sound and can significantly interfere with the passage of ultrasound waves, causing poor image quality or complete signal loss.
Detailed Insights
\item \textbf{Acoustic Impedance Matching}: Human skin and underlying tissues have specific acoustic impedances (the resistance a medium offers to the passage of ultrasound waves). The coupling medium is designed to match the acoustic impedance of the transducer and the skin to minimize the reflection of ultrasound waves at the skin surface.
\item \textbf{Enhanced Contact}: It ensures that the transducer makes good contact with the skin, especially on body parts where the surface is irregular. This uniform contact helps in obtaining consistent and clear ultrasound images.
\end{enumerate}
\endgroup

\noindent \textbf{Practical Example}

\begingroup
\setlength{\parskip}{0pt} 
\setlist[enumerate]{nosep} 
\begin{enumerate}[leftmargin=1cm, label=$\bullet$, itemsep=0pt, parsep=0pt]
\item \textbf{Example of Poor Coupling}: Without the gel, the ultrasound waves might reflect off the skin's surface back into the transducer or scatter in various directions, which would result in a fuzzy or incomplete diagnostic image.
\item \textbf{Example of Good Coupling}: With the gel, ultrasound waves smoothly transition into the body and back to the transducer, providing a clear and detailed view of the internal structures, which is crucial for accurate diagnostics.
\end{enumerate}
\endgroup

In summary, the use of a coupling medium is fundamental in ultrasound diagnostics to enhance the quality of the images obtained, ensuring that the ultrasound waves can be transmitted efficiently into the body and back to the imaging system
~\\

\noindent\textbf{Your preference is \uline{\ \ \ \ } because \uline{\ \ \ \ \ \ \ \ \ \ \ \ \ \ \ \ \ \ \ \ \ \ \ \ \ \ \ \ \ \ }.}
~\\

\noindent\textcolor{gray}{\rule{\linewidth}{2pt}}
~\\

\noindent\textbf{Q9}: Please explain US Doppler imaging. Is the frequency shift always positive?
~\\

\noindent\textbf{Answer A}: Ultrasonic Doppler imaging, or Doppler ultrasonography, is a technique used in medical imaging to visualize and measure the flow of blood or other fluids within the body. This method relies on the Doppler effect, which is the change in frequency of a wave in relation to an observer who is moving relative to the wave source. Here’s how it works in the context of ultrasound:

\begingroup
\setlength{\parskip}{0pt} 
\setlist[enumerate]{nosep} 
\begin{enumerate}[leftmargin=1cm, label=$\bullet$, itemsep=0pt, parsep=0pt]
\item \textbf{Basic Principle}: An ultrasound machine transmits sound waves at a certain frequency into the body. When these waves encounter moving objects, such as red blood cells, the frequency of the reflected waves changes based on the direction and velocity of the object’s motion relative to the ultrasound probe.
\item \textbf{Frequency Shift}: If the object is moving towards the probe, the frequency of the reflected waves increases; if it's moving away, the frequency decreases. This change in frequency is what Doppler ultrasound detects and uses to create images or graphs representing the flow of blood or other fluids.
\end{enumerate}
\endgroup

Is the Doppler Frequency Shift Always Positive? No, the Doppler frequency shift is not always positive:
\begingroup
\setlength{\parskip}{0pt} 
\setlist[enumerate]{nosep} 
\begin{enumerate}[leftmargin=1cm, label=$\bullet$, itemsep=0pt, parsep=0pt]
\item \textbf{Positive Shift}: Occurs when the blood cells (or other targets) are moving towards the ultrasound probe. This results in a higher frequency than the original emitted frequency.
\item \textbf{Negative Shift}: Occurs when the targets are moving away from the probe, resulting in a lower frequency than the transmitted frequency.
The ability to detect both positive and negative shifts is crucial for accurately assessing the direction and speed of blood flow in various medical conditions, such as evaluating blood flow in arteries and veins to detect blockages or abnormalities.
\end{enumerate}
\endgroup
~\\
\noindent\textbf{Answer B}: US Doppler imaging, commonly referred to as Doppler ultrasound, is a diagnostic technique used in medical imaging to evaluate the movement of substances within the body, typically blood. This method extends the capabilities of traditional ultrasound by allowing clinicians to see and measure the velocity and direction of blood flow in vessels, which can help diagnose various conditions, such as blockages, valve defects, and fetal health. Here’s how it works:

\begingroup
\setlength{\parskip}{0pt} 
\setlist[enumerate]{nosep} 
\begin{enumerate}[leftmargin=1cm, label=$\bullet$, itemsep=0pt, parsep=0pt]
\item \textbf{Transmission of Ultrasound Waves}: An ultrasound probe emits high-frequency sound waves into the body.
\item \textbf{Reflection of Sound Waves}: These sound waves bounce off moving red blood cells and other structures within blood vessels.
\item \textbf{Detection of Reflected Waves}: The probe then detects the reflected waves. Due to the motion of the blood cells, these reflected waves will have a frequency that differs from the transmitted waves.
\item \textbf{Doppler Effect}: This change in frequency, known as the Doppler shift, is crucial for analysis. The frequency of the reflected waves increases (positive shift) if the blood cells are moving towards the probe and decreases (negative shift) if they are moving away from the probe.
\end{enumerate}
\endgroup

In Doppler ultrasound, the frequency shift is not always positive. Whether the shift is positive or negative depends on the direction of the flow relative to the ultrasound probe:

\begingroup
\setlength{\parskip}{0pt} 
\setlist[enumerate]{nosep} 
\begin{enumerate}[leftmargin=1cm, label=$\bullet$, itemsep=0pt, parsep=0pt]
\item \textbf{Positive Doppler Shift}: Occurs when the blood cells (or any target) are moving towards the ultrasound probe.
\item \textbf{Negative Doppler Shift}: Occurs when the blood cells are moving away from the ultrasound probe.
\end{enumerate}
\endgroup

The magnitude of the Doppler shift helps in determining the velocity of the blood flow, and the direction of the shift provides information on the direction of flow. This ability to detect the direction and velocity of flow is particularly useful in assessing vascular health and function, as well as in other applications like cardiac valve abnormalities and peripheral vascular disease.
~\\

\noindent\textbf{Your preference is \uline{\ \ \ \ } because \uline{\ \ \ \ \ \ \ \ \ \ \ \ \ \ \ \ \ \ \ \ \ \ \ \ \ \ \ \ \ \ }.}
~\\